\documentclass{vgtc}                          % final (conference style)
\ifpdf%                                % if we use pdflatex
  \pdfoutput=1\relax                   % create PDFs from pdfLaTeX
  \usepackage{graphicx}                % allow us to embed graphics files
  \DeclareGraphicsExtensions{.pdf,.png,.jpg,.jpeg} % for pdflatex we expect .pdf, .png, or .jpg files
\else%                                 % else we use pure latex
  \usepackage{graphicx}                % allow us to embed graphics files
  \DeclareGraphicsExtensions{.eps}     % for pure latex we expect eps files
\fi%

\graphicspath{{figures/}{pictures/}{images/}{./}{vgtc_conference_latex-2019.10.01/}} % where to search for the images

\usepackage{microtype}                 % use micro-typography (slightly more compact, better to read)
\PassOptionsToPackage{warn}{textcomp}  % to address font issues with \textrightarrow
\usepackage{textcomp}                  % use better special symbols
\usepackage{mathptmx}                  % use matching math font
\usepackage{times}                     % we use Times as the main font
\usepackage{cite}                      % needed to automatically sort the references
\usepackage{tabu}                      % only used for the table example
\usepackage{booktabs}                  % only used for the table example
\usepackage{floatrow}
\usepackage{graphicx}

\usepackage{enumitem}
\usepackage{todonotes}

\usepackage{subcaption}
\usepackage{float}
\usepackage{dblfloatfix} 

\usepackage[utf8]{inputenc}

\usepackage{mathtools}

\newenvironment{s_itemize}{
\begin{itemize}[leftmargin=*]
  \setlength{\itemsep}{3pt}
  \setlength{\parskip}{0pt}
  \setlength{\parsep}{0pt}
}{\end{itemize}}

\newenvironment{s_enumerate}{
\begin{enumerate}[wide, labelwidth=!, labelindent=0pt]
  \setlength{\itemsep}{2pt}
  \setlength{\parskip}{0pt}
  \setlength{\parsep}{0pt}
}{\end{enumerate}}

\usepackage{amsfonts} 

\usepackage{pifont}% http://ctan.org/pkg/pifont
\usepackage{amsmath}

\let\oldhat\hat 
\renewcommand{\hat}[1]{\oldhat{\mathbf{#1}}}

\usepackage{xcolor}

\newcommand{\changetextxx}[1]{\textcolor{black}{#1}}
\newcommand{\changetextt}[1]{\textcolor{black}{#1}}

\usepackage{rotating}

\usepackage{floatrow}
\onlineid{3861}

\vgtccategory{Research}

\title{Towards Open-World Gesture Recognition}

\author{Junxiao Shen\thanks{e-mail: junxiao.shen@bristol.ac.uk}\\ %
        \scriptsize Meta Reality Labs Research \\ \scriptsize University of Bristol %
\and Matthias De Lange\\ %
     \scriptsize Meta Reality Labs Research %
\and Xuhai Xu\\ %
     \scriptsize Meta Reality Labs Research %
\and Enmin Zhou\\ %
     \scriptsize Meta Reality Labs Research %
\and Ran Tan\\ %
     \scriptsize Meta Reality Labs Research %
\and Naveen Suda\\ %
     \scriptsize Meta Reality Labs Research %
\and Maciej Lazarewicz\\ %
     \scriptsize Meta Reality Labs Research
\and Per Ola Kristensson\\ %
     \parbox{1.4in}{\scriptsize \centering University of Cambridge} %
\and Amy Karlson\\ %
     \scriptsize Meta Reality Labs Research %
\and Evan Strasnick\\ %
     \scriptsize Meta Reality Labs Research}

\teaser{
  \centering
  \includegraphics[width=\linewidth]{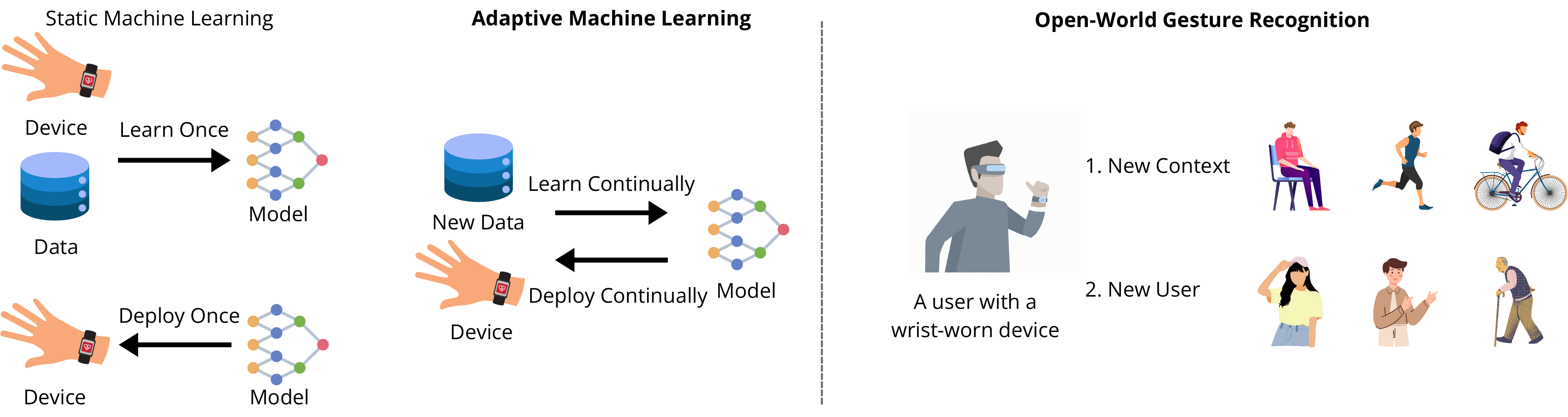}
    \caption{\textbf{Static Machine Learning}: The model undergoes a single training session using existing data and is subsequently deployed to the device. 
    \textbf{Adaptive Machine Learning}: Recognizing the dynamic nature of real-world scenarios, this approach continuously updates and refines the model as new data become available. Consequently, updated models are regularly deployed to the device.
    \textbf{Open-World Gesture Recognition}: In the context of wrist-worn devices utilized for gesture recognition, there are two distinct cases where new data will appear. In these cases, the gesture recognition model must adapt and learn from this new information. }
    \label{fig:teaser2}
  \label{fig:teaser}
}

\abstract{Providing users with accurate gestural interfaces, such as gesture recognition based on wrist-worn devices, is a key challenge in mixed reality. However, static machine learning processes in gesture recognition assume that training and test data come from the same underlying distribution. Unfortunately, in real-world applications involving gesture recognition, such as gesture recognition based on wrist-worn devices, the data distribution may change over time. We formulate this problem of adapting recognition models to new tasks, where new data patterns emerge, as \emph{open-world gesture recognition} (OWGR). We propose the use of continual learning to enable machine learning models to be adaptive to new tasks without degrading performance on previously learned tasks. However, the process of exploring parameters for questions around when, and how, to train and deploy recognition models requires resource-intensive user studies may be impractical.
To address this challenge, we propose a design engineering approach that enables offline analysis on a collected large-scale dataset by systematically examining various parameters and comparing different continual learning methods. Finally, we provide design guidelines to enhance the development of an open-world wrist-worn gesture recognition process.
} % end of abstract

\CCScatlist{
  \CCScatTwelve{Computing methodologies}{Machine learning}{Learning paradigms}{Lifelong machine learning};
  \CCScatTwelve{Human-centered computing}{Human computer interaction (HCI)}{HCI design and evaluation methods}{User models}
}

\begin{document}


\firstsection{Introduction}
\label{sec:introduction}

\maketitle
% \section{Introduction} %for journal use above \firstsection{..} instead

% \section{Introduction} %for journal use above \firstsection{..} instead
Wrist-worn gesture recognition using inertial measurement unit (IMU) signals offers a convenient, always-on interface for various applications in mixed reality~\cite{jiang2017feasibility,siddiqui2020multimodal,reifinger2007static,sagayam2017hand}.
Currently, most hand gesture recognition algorithms are optimized for \emph{closed-world} settings, where training and test data come from the same underlying distribution~\cite{SHI2021183}.
However, in real-world applications~\cite{bendale2015towards,ren2022reinforcement}, the assumption that training and test data belong to the same underlying distribution no longer holds as: (1) new gesture data continuously arrives with changing characteristics; (2) gesture data may change over time; and (3) entirely new data patterns can emerge.
We here call this problem \emph{open-world gesture recognition} (OWGR).
A static machine learning process, in which a recognition model is trained once and subsequently deployed, cannot effectively tackle this problem. What we seek is an \emph{adaptive} machine learning process that can continuously train and deploy the model on newly \emph{emerging} data, as illustrated in Figure~\ref{fig:teaser2}.

Various approaches can be used for adaptive machine learning. Simply retraining a recognition model with the entire joint dataset, past data and new data from new tasks, is sometimes infeasible due to limited computational power on embedded devices. 
Storing the entire past dataset is also challenging due to the limited memory of the device and privacy concerns.
On the other hand, if a model trained from past data is na\"{i}vely fine-tuned on a new task, that model will dramatically decrease the recognition performance on the old tasks~\cite{saponas_enabling_2009}.
This is called catastrophic forgetting.
Continual learning, sometimes called lifelong learning, methods are specifically designed to alleviate catastrophic forgetting by balancing the trade-off between plasticity (transferring knowledge from an old task to a new task) and stability (catastrophic forgetting).

We identify two real-world cases that fall under OWGR, \emph{new context} and \emph{new user}, as shown in Figure~\ref{fig:teaser2}.
% The first case is \emph{new context}, where the gesture recognition model adaptively learns to recognize existing gestures in new situations, such as performing a pinch action when \textit{walking} and \textit{running}, without forgetting old contexts such, as performing a pinch action when \textit{standing}. 
% The second case is \emph{new user}, where the recognition model learns to recognize existing gestures as performed by new users, such as when a device is shared between users, or a newly-bought device calibrates to a user for the first time. 
In all two cases, we wish to preserve performance on previously learned tasks, to avoid catastrophic forgetting, while at the same time learning the new task.
Here, a \emph{new task} may refer to a new context, or a new user. 
In Section~\ref{sec:owgr}, we present a key contribution in this paper---a detailed formulation of the \emph{open-world gesture recognition} problem and its associated real-world scenarios, in terms of these two cases: (1) new context; and (2) new user.

\begin{figure}[t]
    \centering
    \includegraphics[width=0.85\textwidth]{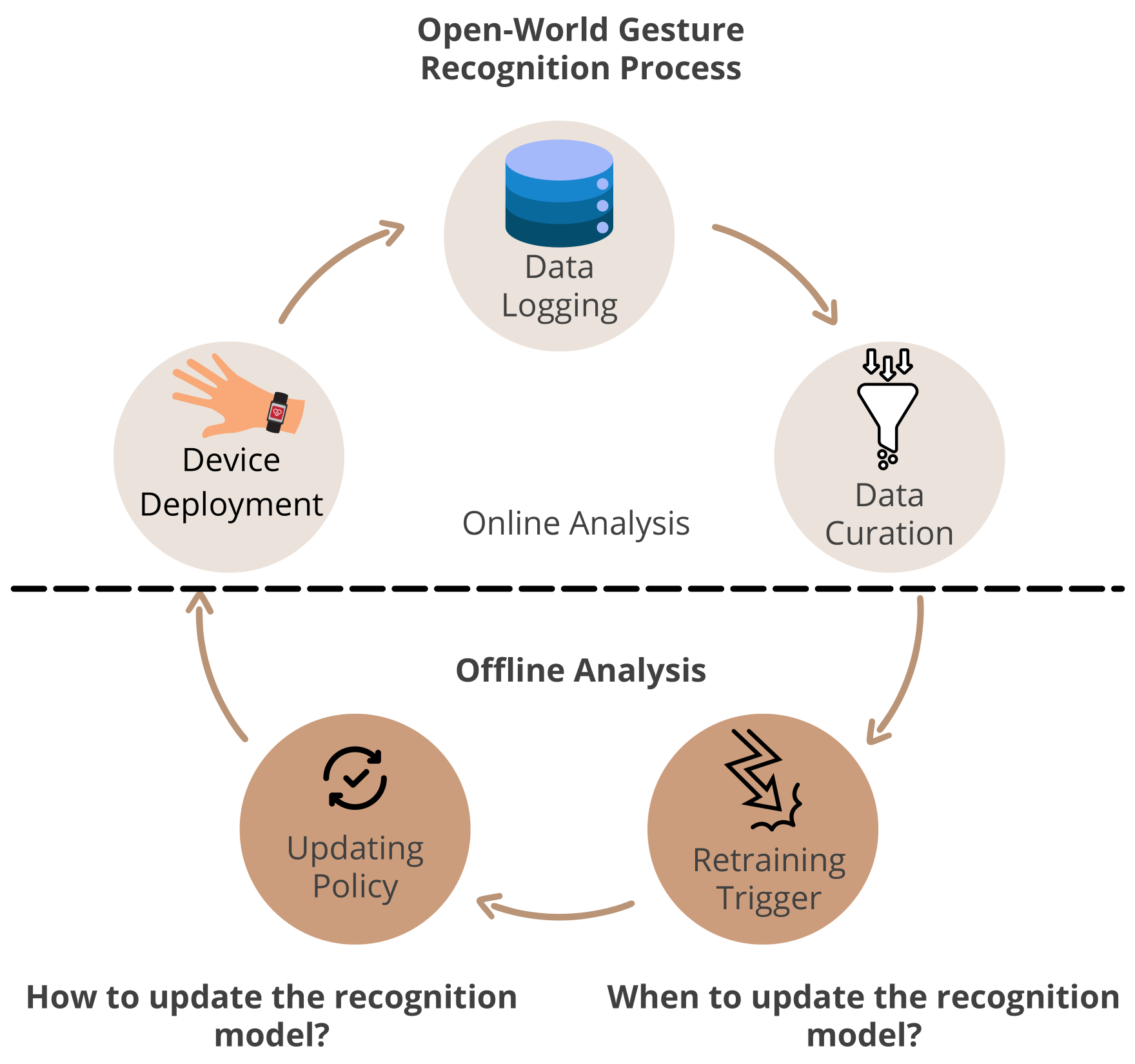}
    \caption{The \emph{open-world gesture recognition} process is structured around five stages. Out of these, two stages including \textit{retraining trigger} and \textit{updating policy} introduce essential design considerations, all underpinned by an engineering approach that facilitates offline analysis.  In contrast, the other three stages including \textit{device deployment}, \textit{data logging} and \textit{data curation} require online analysis. }
    \label{fig:process}
\end{figure}

In a static machine learning process, we only need to log the data from the device, optimize the recognition model on the logged data to best fit the data once, and then deploy a trained model onto the device. 
However, in an adaptive machine learning process for OWGR, we additionally introduce additional stages as illustrated in Figure~\ref{fig:process}.
In particular, this OWGR process includes two additional stages that require careful design considerations: (1) \emph{when} to update the model; and (2) \emph{how} to update the model.
When refining a wrist-worn \emph{open-world gesture recognition} system, developers face challenges due to the complexity of user studies and numerous parameters. For instance, evaluating \textit{new context} requires recruiting participants to perform gestures in various real-world contexts, not just in a lab. Additionally, optimizing the OWGR process often needs multiple evaluations, making it difficult to execute and validate.

% When a developer of a wrist-worn \emph{open-world gesture recognition} system seeks to assess and refine their approach, particularly when delving into the aforementioned two cases, challenges emerge. These stem from the intricacies of conducting actual user studies, given the numerous parameters that have to be taken into consideration.
% For instance, evaluating an OWGR process under \textit{new context} via a user study could necessitate recruiting representative participants to perform gestures across many distinct contexts in the real world, as opposed to in a lab. Further, an optimization of the OWGR process might require multiple such evaluations, which makes the process highly challenging to carry out and validate.

\changetextt{
We therefore follow a prior approach~\cite{Kristensson2020Design}, which proposes using a design engineering approach to assess a process that is challenging to evaluate via user studies. We have adopted this design engineering approach by translating it into a strategy that outlines a functional design of \emph{open-world gesture recognition} and examines its conceptual design in which the key function of the process (such as \texttt{Updating Policy}) is translated into six potential function carriers (five different implementations of continual learning methods and one finetuning baseline). We then distinguish between controllable and uncontrollable parameters of these functions and assess the process's effectiveness through quantitative envelope analysis.
Moreover, given the lack of prior research that focuses specifically on OWGR, it is crucial to first apply the most advanced and generalizable continual learning methods to address this issue. Only after extensive exploration and evaluation of OWGR using other more generalizable and well-established continual learning methods are we in a position to best understand how to propose new algorithms to further enhance the system. Therefore, proposing a new algorithm is beyond the scope of this paper.
}

\changetextt{
Such a design engineering approach allows us to assess controllable and uncontrollable parameters offline, thus sidestepping the need for on-ground deployments and tests with real users in the first instance.
To this end, we first collect a large-scale dataset using the IMU sensor from wrist-worn devices. To capture diversity in the aforementioned two real-world cases, the dataset contains 50 users performing four different gestures in 25 different contexts. 
We then construct a surrogate data model and a surrogate task model, identify their controllable and uncontrollable parameters, and subsequently perform an envelope analysis by varying these parameters, which allows us to assess emergent qualities of the system as a whole via simulation.
This approach is then able to illustrate that using a continual learning method for \emph{open-world gesture recognition} significantly improves gesture recognition accuracy and reduces catastrophic forgetting compared to a finetuning method. 
This forms our second contribution to the paper. 
}

\changetextt{
Designers using engineering methodologies to refine the OWGR process face limitations in data collection and computational resources. Envelope analysis, being computationally intensive, demands extensive simulations and parameter sweeps. Our study required over 40,000 GPU hours for dataset collection and offline evaluation of continual learning methods, highlighting the challenge for resource-constrained designers.
To aid these designers, we provide essential design guidelines derived from our methodology. Our diverse dataset, covering various users and contexts, ensures the generalizability of these guidelines. Therefore, our third contribution is offering distilled guidelines for developing OWGR processes based on our engineering approach using envelope analysis.
}

% Despite employing a design engineering methodology to assess and refine the OWGR process, designers are often restricted by limited abilities to collect large-scale data and limited availability of computational resources. This is because envelope analysis remains computationally demanding due to its intensive simulation nature, necessitating parameter sweeping. 
% In this paper, the collection of an extensive dataset and our subsequent offline examinations, which encompasses the training and evaluation of different continual learning methods, imposes a significant resource burden, consuming over 40,000 GPU hours.
% Consequently, designers with constrained resources will find it challenging to optimize the OWGR process. Therefore, we also distill essential design guidelines based on our engineering methodology's outcomes to guide designers working on OWGR.
% Our assembled dataset exhibits generalizability due to its substantial diversity in terms of users and contexts. This in combination with our use of representative continual learning methods, allows the design guidelines presented in this paper to be widely applicable.
% Hence, this paper's third contribution is the provision of distilled design guidelines for developing OWGR processes, underpinned by our design engineering approach using envelope analysis.

In summary, this paper makes three contributions: (1) a detailed formulation of the \emph{open-world gesture recognition} (OWGR) problem; (2) a design engineering approach for allowing \emph{open-world gesture recognition} to be tractable in practice; and (3) design guidelines for developing OWGR processes.

% ~\footnote{We plan to release the code and dataset as a public research benchmark for OWGR in conjunction with publishing the paper.}

\section{Related Work}
% In this section, we discuss the related work.

\subsection{Wrist-Worn Devices with Mixed Reality}
% Wrist-worn devices have been naturally integrated with mixed reality devices~\cite{anthes2016state,feiner2002augmented}.
Gestural interaction offers an intuitive and natural method for interfacing with mixed reality~\cite{yang2019gesture}. More commonly, the front-facing cameras of Head Mounted Displays (HMDs) track the hands, enabling hand gestures for text entry, virtual object manipulation, and menu interaction, such as dragging virtual items or making selections~\cite{buckingham2021hand}. However, a fundamental drawback of using front-facing cameras is the necessity for the hands to remain within the camera's view. This means users cannot perform gestures with their hands under the table or in their pockets. Once hands leave the camera's tracking region, gestural interaction becomes impossible. 

To overcome this limitation, wrist-worn devices have been employed for gestural interaction with mixed reality devices~\cite{anthes2016state,feiner2002augmented}. Gesture recognition is conducted using these wrist-worn devices, eliminating the need for HMD cameras to detect hands and gestures. Using wrist-worn devices facilitates gestural interactions even when hands are under the table or in a pocket, offering a seamless user experience. Moreover, wrist-worn devices can recognize gestures in a broader variety of ways compared to front-facing cameras, which primarily rely on computer vision techniques.
There has been extensive research exploring hand gesture recognition using a wide range of wrist-worn sensing modalities, such as RGB cameras~\cite{wu_back-hand-pose_2020,hu_fingertrak_2020,xu_hand_2018}, infrared (IR) ranging~\cite{mcintosh_sensir_2017},
inertial measurement unit (IMU)~\cite{kim_imu_2019,escalera_gesture_2017},
acoustics~\cite{nandakumar_fingerio_2016,laput_sensing_2019}, electromyography (EMG)~\cite{saponas_enabling_2009,caramiaux_understanding_2015,mcintosh2016empress}, electrical impedance tomography~\cite{zhang_tomo_2015}, pressure~\cite{dementyev_wristflex_2014,jung_wearable_2015}, radar~\cite{lien_soli_2016}, stretch sensors~\cite{strohmeier_flick_2012}, magnetic sensors~\cite{chen_finexus_2016,parizi_auraring_2019}, and bio-capacitive effects~\cite{truong_capband_2018}.
IMU sensors are widely used for gesture recognition owing to their low cost and low power characteristics~\cite{Zhang2018Application,Xu2022Enabling}. 

\subsection{Gesture Recognition}
\label{sec:wrist_worn_gesture}

% In this work, we focused on the IMU sensors because of their ubiquity and potential for generalizability.

As for gesture recognition techniques, early trajectory-based gesture recognition methods (e.g., dynamic time warping (DTW)~\cite{liu_uwave_2009} and hidden Markov models (HMM)~\cite{mckenna_comparison_2004}) can recognize simple gesture trajectories such as drawing a shape ~\cite{mckenna_comparison_2004}).
However, these methods do not work well for more complex and fine-grained gestures such as pinching or making a fist.
Recently, researchers and practitioners mainly use data-driven approaches by collecting a labeled gesture dataset. They then either train traditional models such as SVM, trees, (e.g., \cite{georgi_recognizing_2015,iravantchi_beamband_2019}) or deep learning models when the dataset is large enough (e.g., \cite{hu_fingertrak_2020,yeo_opisthenar_2019}).
However, most prior work focuses on recognizing a pre-defined gesture set. There are very few works addressing \emph{open-world gesture recognition}.
Xu \textit{et al.} proposed a few-shot learning framework for gesture authoring~\cite{Xu2022Enabling}. 
Shen \textit{et al.} proposed a deep-learning model that can learn a new gesture with a synthetic dataset that is generated from a few data samples with a deep generative model~\cite{Shen2022Gesture,Shen2021Imaginative,Shen2021Simulating}. Wang \textit{et al.} ~\cite{wang2020catnet} identified a continual learning application for lifelong egocentric gesture recognition such that a VR system allows users to customize gestures incrementally. 
While these approaches are useful for models to learn new gestures, they are not appropriate for the two cases we identified, \emph{new context} and \emph{new user}. Moreover, Xu \textit{et al.}~\cite{Xu2022Enabling} and Shen \textit{et al.}~\cite{Shen2022Gesture} also fail to address the catastrophic forgetting problem when old gesture data are revisited.

% Wang \textit{et al.} ~\cite{wang2020catnet} identify a continual learning application for lifelong egocentric gesture recognition such that a VR system allows users to customize gestures incrementally. Our work differs in several ways. First, they only applied one continual learning method, iCaRl~\cite{rebuffi2017icarl}. 
% iCaRl requires large amounts of memory, which can be infeasible on wrist-worn devices. In contrast, we apply five different continual learning methods with varying trade-offs in our work. 
% Second, they did not apply continual learning to examine our proposed two cases: \emph{new context} and \emph{new user}.
% Second, their application case is under the assumption of \textit{class-incremental learning} where task identity needs to be inferred and we follow the \textit{task-incremental learning} setting where task identity is given~\cite{van2019three}. 
% Third, they focused on egocentric gesture recognition with RGB and depth images as inputs, whereas we focus on wrist-worn gesture recognition with inputs being time-series IMU data. 
% Lastly, we provide a systematic evaluation on various continual learning methods for different application scenarios, aiming to provide general design guidelines.

\subsection{Design Engineering} 
Design engineering is a holistic method to create products and systems. This approach transitions from the initial stage of problem identification to tackling design-related aspects during the life cycle of a product or system, including production, upkeep, and eventual decommissioning. Validating certain systems, such as a a context-aware sentence retrieval system for AAC users, is extremely challenging~\cite{Kristensson2020Design} as such systems require tailored setups and extended use before benefits emerge. Asking an AAC user to switch devices, potentially waiting months for improved communication, raises both logistical and ethical concerns.
Prior work~\cite{Kristensson2020Design} has used a design engineering approach to tackle this problem. This approach entails defining controllable and uncontrollable parameters for the design and examining the potential effectiveness of such systems through quantitative envelope analysis by varying the parameters through simulation~\cite{Kristensson2020Design}.
Such a design engineering method reveals insights about the feasibility of such systems without the immediate need to develop, introduce, and observe systems for extended time periods. In addition, it allows examination of emergent qualities of systems that would be very difficult to study in empirical user studies, such as mechanisms explaining users strategies~\cite{kristensson2021design}.
Similarly, Shen \textit{et al.}~\cite{Shen2022KWickChat} also used a design engineering approach to evaluate a multi-turn dialogue system for AAC using context-aware sentence generation by bag-of-keywords.
The design engineering approach we propose in this paper shares a similar philosophy with this prior work work.

\subsection{Continual Learning}
\label{sec:continual_learning}

% In a continual learning setup, the learning system is exposed to a series of tasks, sequentially, over time. Each task is represented by a specific data distribution. The tasks can be related or completely different from each other.
% The new tasks can come from various domains depending on the application, such as new classes in image classification, new users in recommendation systems, or new environments in reinforcement learning. The model is expected to learn these new tasks while maintaining its performance on the old tasks.
% One of the key aspects of the continual learning setup is that the model typically does not have access to the data from previous tasks while learning a new one, due to memory constraints or privacy issues.

In a continual learning setup, the learning system is exposed to a sequence of tasks over time, each represented by a specific data distribution. The tasks may be related or completely different. The model is expected to learn new tasks while maintaining performance on previous ones, typically without access to data from earlier tasks due to memory constraints or privacy issues.
When a model, previously trained on older tasks, is simply fine-tuned on new tasks, its accuracy on the old tasks rapidly deteriorates.
It loses its proficiency in tasks it had previously mastered, effectively replacing the old knowledge with the new. 
This phenomenon is referred to as catastrophic forgetting. This problem occurs because conventional neural networks update their weights primarily based on the most recent data during training, a process that can result in the erasure of previously acquired patterns and knowledge.
More specifically, gradient-based optimization algorithms prioritize minimizing the loss of the current training task, often disregarding previous task parameter settings~\cite{french1999catastrophic}. 
% Although constraining parameter updates can alleviate this issue, it hampers the model's ability to effectively learn new tasks. 
% This creates a dilemma between stability (retaining knowledge of old tasks) and plasticity (learning new tasks), posing a challenge for continual learning~\cite{french1999catastrophic}. 

To address this challenge, continual learning methods aim to leverage previously seen information during training to improve performance on new tasks. The objective is to overcome the forgetting of learned tasks (stability) and utilize prior knowledge to achieve better performance and faster convergence on new tasks (plasticity)~\cite{delange2021continual}. There are three prominent families of methods for continual learning: replay methods, regularization-based methods, and parameter isolation methods~\cite{delange2021continual}. 
Replay methods~\cite{buzzega2021rethinking,pritzel2017neural,rebuffi2017icarl,isele2018selective,chaudhry2019continual} use rehearsal to mitigate catastrophic forgetting, storing examples from old tasks in memory to be replayed throughout incremental task training.
However, these methods are less memory efficient.
Regularization-based methods introduce a regularization term in the loss function of the model~\cite{mallya2018piggyback, riemer2018learning,aljundi2018memory,zenke2017continual}.
Parameter isolation methods either dedicate different subsets of the model parameters to each task to prevent any possible forgetting or expand the size of the model to acquire new knowledge from new tasks~\cite{rusu2016progressive,mallya2018packnet,yoon2017lifelong}. Neither regularization-based nor parameter isolation methods store past data in memory.

There are also other different learning paradigms, such as multi-task learning~\cite{zhang2021survey}, transfer learning~\cite{weiss2016survey}, meta learning~\cite{vanschoren2018meta} and online learning~\cite{shalev2012online}. 
However, while most of these methods focus on plasticity, they usually fail to address catastrophic forgetting, leading to a dramatic decrease in the performance with old tasks~\cite{delange2021continual}. \changetextxx{Being able to avoid catastrophic forgetting is critical in OWGR as it is not solely about quickly learning new data - a challenge that can be addressed by few-shot learning~\cite{Sung2017LearningTC} or domain adaptation~\cite{Mansour2009DomainAL} - but more importantly, the core challenge of OWGR is to retain old data. More specifically, domain adaptation focuses on adapting to new domains without necessarily retaining performance on the original domain, which does not directly address catastrophic forgetting. Few-shot learning deals with learning new tasks or adapting to new data quickly from few examples, but it does not inherently address learning multiple tasks sequentially without forgetting. These two methods are specifically designed to optimize plasticity (transferring knowledge from an old task to a new task), without considering stability (catastrophic forgetting). Therefore, these learning paradigms are not well-suited to the OWGR problem. Hence, we do not directly compare continual learning methods with other learning paradigms in this paper. }

\section{Open-World Gesture Recognition (OWGR)}
\label{sec:owgr}

% We start by characterizing the OWGR by two real-world cases in detail. We then formally formulate our problem.

\subsection{Real-World Cases}
\label{sec:cases}
% We identify two general real-world cases for OWGR

\begin{s_enumerate}
\item \textbf{\emph{New context}} is where each task represents different environmental/activity contexts, such as \textit{standing, walking, riding in a car, pushing stroller/cart, laying down} etc.
Therefore, we desire an OWGR system that is initialized on training data only from \textit{standing} contexts to be able to incrementally learn on new contexts such as \textit{walking} and \textit{riding in a car}, while also preserving the knowledge from the previous \textit{standing} contexts. 
These contexts can be further split into more detailed contexts.
For example, \textit{standing} includes \textit{standing with hand up to the chest level} and \textit{standing with hand hanging}, and \textit{walking} includes \textit{walking with hand up to the chest level} and \textit{walking with hand hanging}.
Our dataset includes this finer granularity of activity context.
In practice, we need a \textit{task descriptor} to inform the model of each task in the \emph{new context}. In the real-world deployment, the description can be determined by the contextual information provided by other devices, such as location or an HMD.
\item \textbf{\emph{New User}} is where each new task is a separate user of the device.
A shared wrist-worn device should quickly adapt to new users, as we assume the behaviors of performing gestures from different users are different. Moreover, we also desire the device to not forget the old users as this shareable device may be switched back to old users.
The task descriptor is reflected by the user's identity.
\end{s_enumerate}

\subsection{Problem Formulation}
\label{sec:formulation}

We formulate the OWGR problem more formally in this section.
In a \emph{task-incremental setting} for \emph{open-world gesture recognition}, the tasks (new context/gesture/user) arrive sequentially and the learner optimizes until convergence within each task \cite{rusu2016progressive,fernando2017pathnet,rebuffi2017icarl}.
This setting has the following assumptions:
\begin{s_enumerate}
    \item Each new context/gesture/user would have a batch of data points. Therefore, the gesture data arrives sequentially in batches, with each batch corresponding to one task. The continual learning takes one batch at a time, while we can still perform offline learning \textit{within} each task. % Some methods will train the data in each task until convergence.
    \item The recognition model will have a multi-head configuration, as each task needs a separate output layer. Therefore, a task descriptor is also fed into the recognition model since the algorithm needs to know which head to use for that specific task. In practice, this task descriptor would be generated through various sensing modalities (e.g., activity recognition and location information~\cite{vrigkas2015review}). 
    For example, a classifier using egocentric video from a head-mounted display would estimate an activity context. 
    We discuss the feasibility of this method in Section~\ref{sec:when} and Section~\ref{sec:limitations}.
    \changetextxx{
    Despite a multi-head configuration may not be necessary for the \emph{new user} case, the multi-head approach simplifies the comparative analysis of different continual learning methods --- allowing for consistent evaluation with minimal adjustments.}
\end{s_enumerate}

Traditional gesture recognition assumes testing and training data share similar characteristics, that is, they are independent and identically distributed (i.i.d). For this problem setting, we represent the training set as $D_{train} = {(x_{i},y_{i})}^{n}_{i=1}$, where $n$ denotes the number of training data available, contains a feature vector $x_{i}\in X$, and a target vector $y_{i}\in Y$. 
Each data pair in the training set $(x_{i},y_{i})$ is sampled from an i.i.d probability distribution, which corresponds to a single task. 
The goal is to minimize the empirical risk of all data in the task by optimizing the parameters $\theta$ of the gesture recognition model~\cite{vapnik1999overview}:
\begin{equation}
     \frac{1}{\left | D_{train} \right |}\sum_{(x_{i},y_{i})\in D_{train}}^{} \mathcal{L}(f(x_{i}; \theta),y_{i}),
    \label{eq:risk1}
\end{equation}
where the loss function $\mathcal{L}$ penalizes prediction errors, and
$f$ denotes the trained gesture recognition model.

However, the i.i.d assumption no longer holds in OWGR, where data arrives in an online fashion with changing characteristics. 
% A more realistic assumption is that data is not i.i.d with respect to any fixed probability distribution. 
A learner will observe the continuum of data as a triplet $(x_{i},y_{i},t_{i})$. $t_{i} \in T$ is a task descriptor identifying the task associated with the pair $(x_{i},y_{i})$. The pair is sampled from a probability distribution $P_{t_{i}}$ corresponding to the task descriptor $t_{i}$. 
In each task $t_{i}$, the gesture data pair $(x_{i},y_{i})$ is locally i.i.d.
% This means for each gesture data pair $(x_{i},y_{i})$ in the task described by $t_{i}$, if the task descriptor is the same, the gesture data pair is locally i.i.d. with tasks arriving in sequence. 

The goal of continual learning is to minimize the statistical risk by optimizing the parameters $\theta$ of the gesture recognition model $f$~\cite{delange2021continual}: $
    \sum_{t=1}^{\tau }\mathbb{E}_{X^{(t)},Y^{(t)}}\left [ \mathcal{L}(f_{t}(X^{(t)}; \theta),Y^{(t)}) \right ]
$, with Expectation $\mathbb{E}$ of the loss function denoted by $\mathcal{L}$, the number of tasks seen so far denoted by $\tau$, $X^{(t)}$ being a set of data samples for task $t$, $Y^{(t)}$ being the labels correspondingly, and $f_{t}$ representing the recognition model for task $t$. For the current task $\tau$, the statistical risk can be approximated by the empirical risk~\cite{delange2021continual}: $    \frac{1}{N_{\tau}}\sum_{i=1}^{N_{\tau}}\left [ \mathcal{L}(f_{\tau}(x_{i}^{(\tau)}; \theta),y_{i}^{(\tau)}) \right ]$.

\subsection{Large-Scale Data Collection}
\label{sec:data_collection}
We collected a large-scale dataset of inertial measurement unit (IMU) data consisting of 6 dimensions (3-axis accelerometer and 3-axis gyroscope). \footnote{Figure 1 in the Appendix illustrates exemplary IMU signals collected.}

\begin{figure}[t]
  \centering
  \hspace*{\fill}  % Create equal horizontal space on the left
  \begin{subfigure}[b]{0.23\linewidth}
    \centering
    \includegraphics[width=0.5\linewidth]{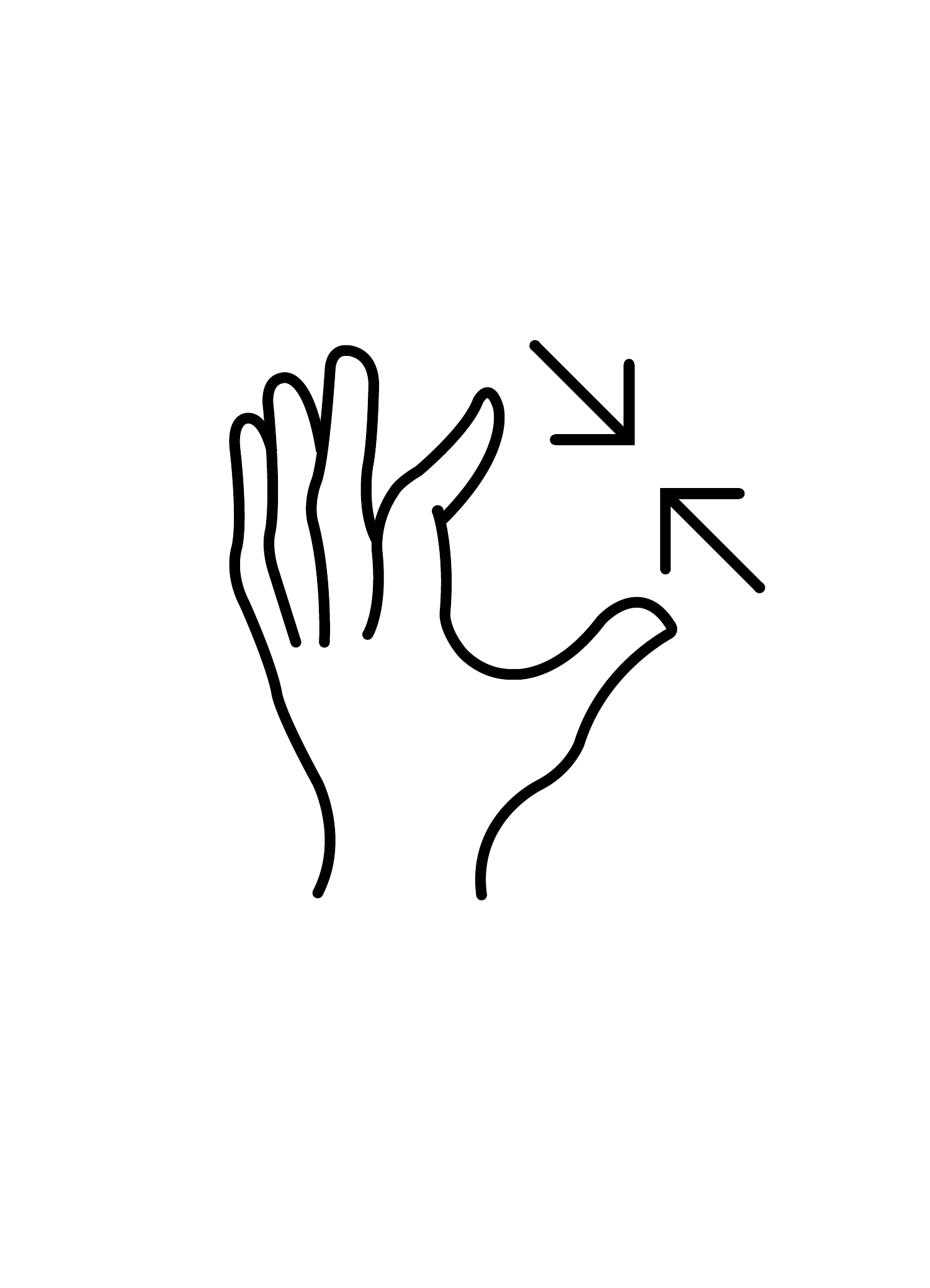}
    \caption{Single Pinch}
  \end{subfigure}
  \hfill  % Creates horizontal space to evenly distribute the subfigures
  \begin{subfigure}[b]{0.23\linewidth}
    \centering
    \includegraphics[width=0.5\linewidth]{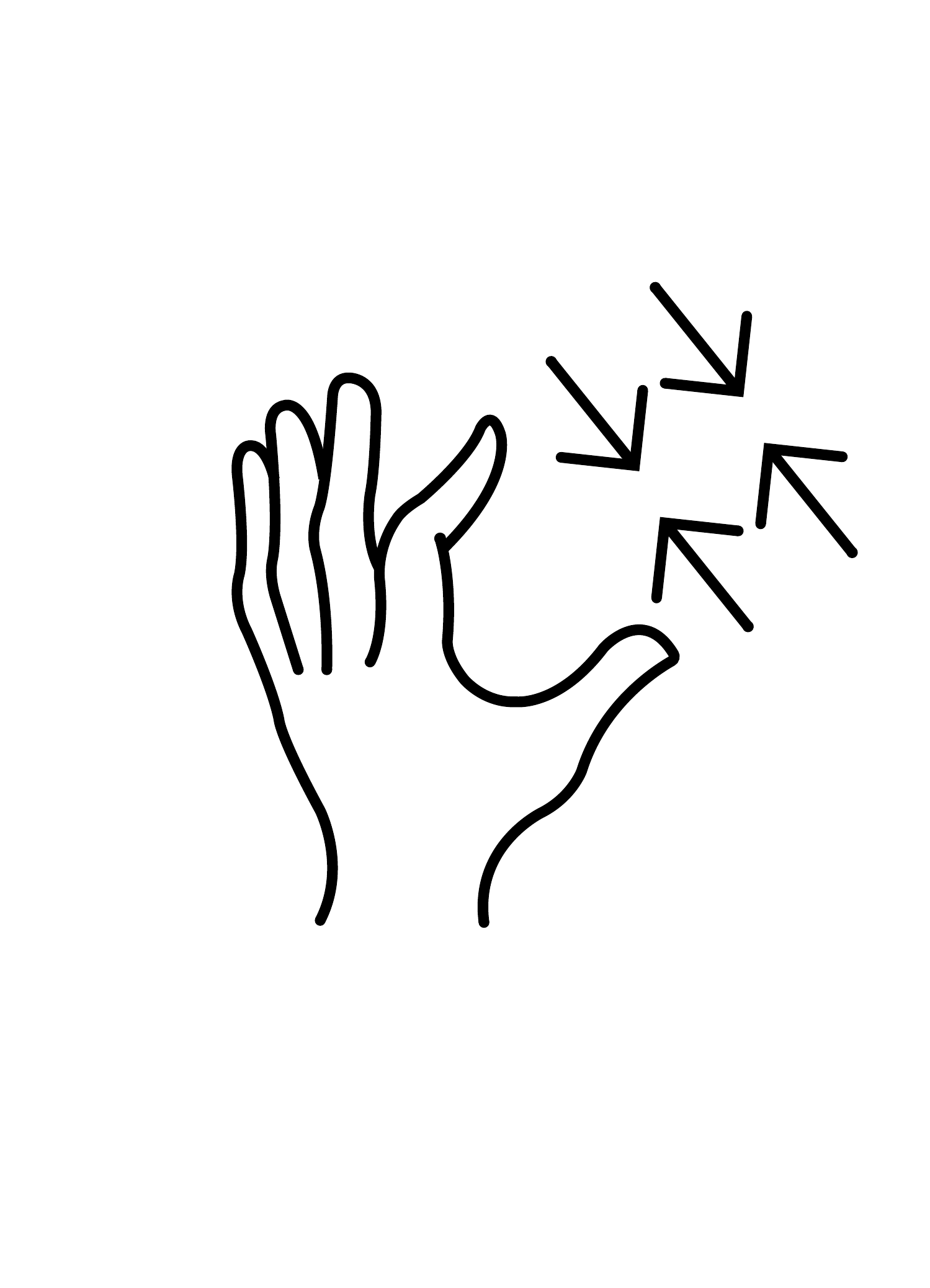}
    \caption{Double Pinch}
  \end{subfigure}
  \hfill
  \begin{subfigure}[b]{0.23\linewidth}
    \centering
    \includegraphics[width=0.5\linewidth]{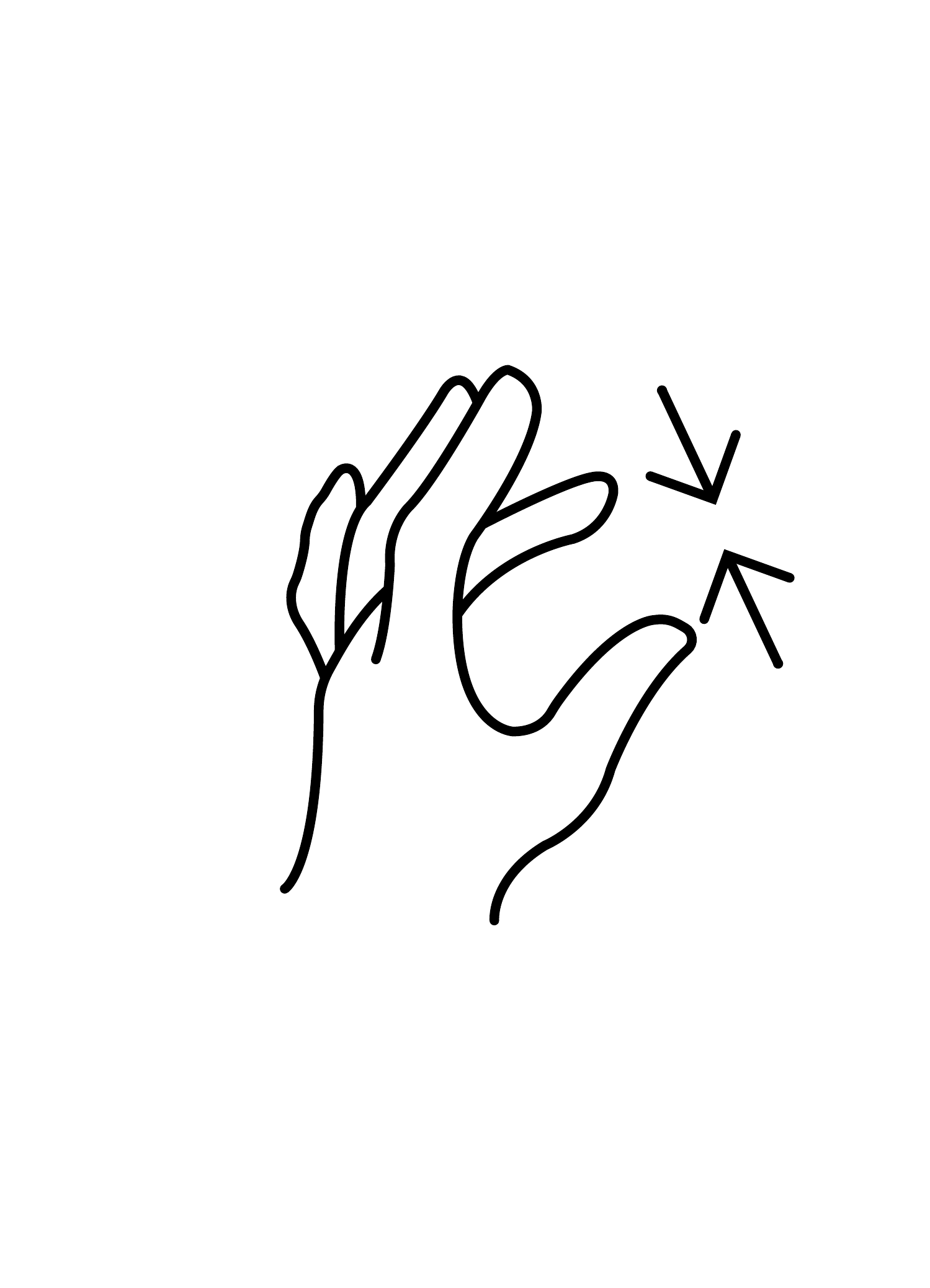}
    \caption{Middle Pinch}
  \end{subfigure}
  \hfill
  \begin{subfigure}[b]{0.23\linewidth}
    \centering
    \includegraphics[width=0.5\linewidth]{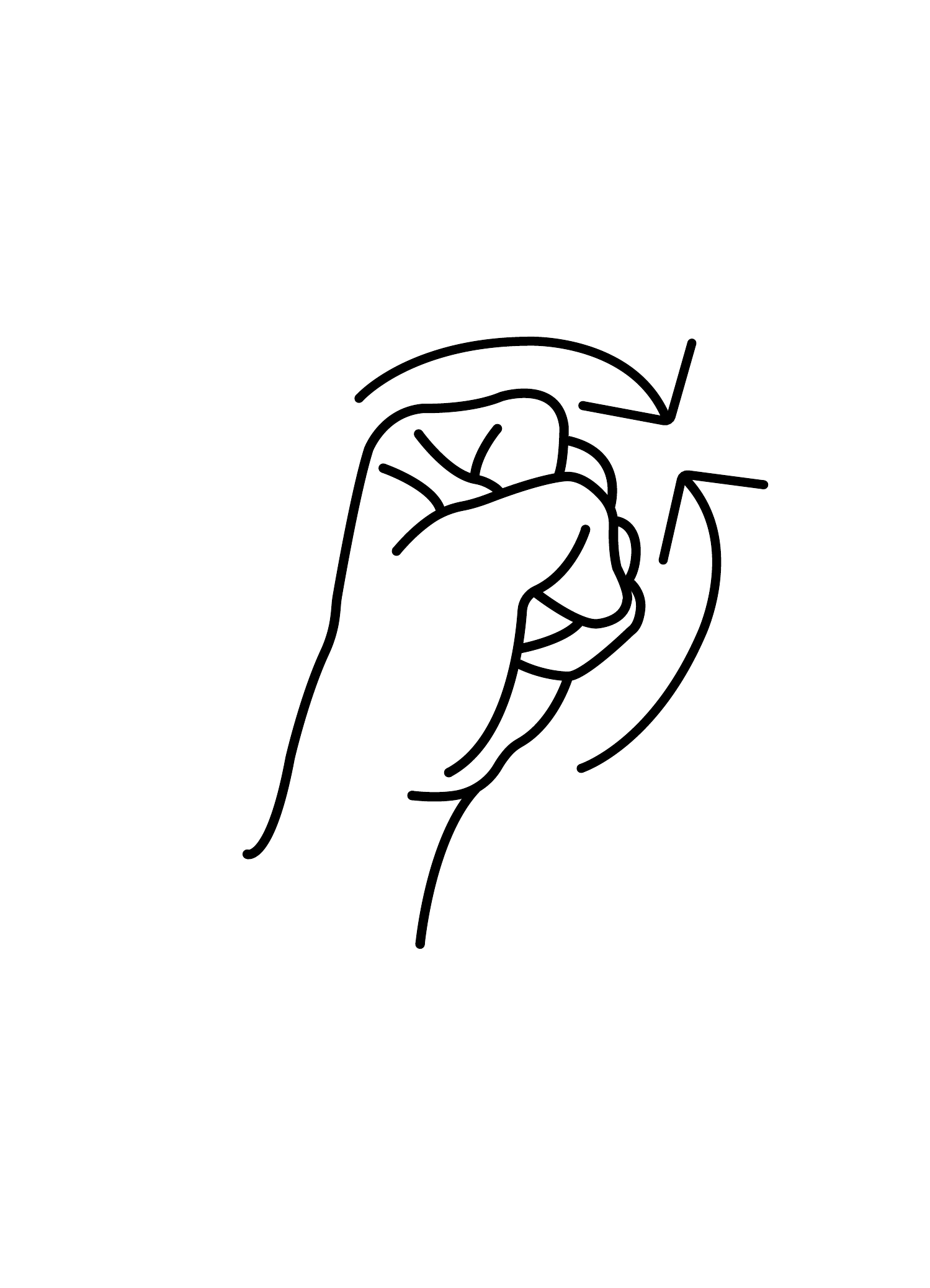}
    \caption{Fist Clench}
  \end{subfigure}
  \hspace*{\fill}  % Create equal horizontal space on the right
  \caption{Four dynamic gestures.}
  \label{fig:gestures}
\end{figure}

\begin{s_enumerate}
    \item \textbf{Gestures}: We collected data from four dynamic gestures: single pinch, double pinch, middle pinch, and fist clench (see Figure~\ref{fig:gestures}). 
    The specific set of gestures was chosen for their distinguishability, and their representation of common hand actions. This selection ensures familiarity, versatility, and intuitive interaction for users. Apple's AssistiveTouch gestures are also similar to the gesture set.

    \item \textbf{Participants and Apparatus}: 50 participants were recruited (24 self-identified female, 26 male) with a wide coverage of age range (min = 18, max = 61, mean = 35).
    The majority of the users were right-handed (N=44).
    We collected data using a wrist-worn watch-like device equipped with IMU sensors. Initially, we gathered raw data at a high sampling rate of 800 Hz to maximize our dataset's potential. 
    However, during model training/testing, we found that a sampling rate of 100 Hz was sufficient. 
    Therefore, for the rest of the paper, we exclusively uses 100 Hz data.
    This rate represents a balanced compromise between a high sampling rate, which would lead to increased power consumption, and a low sampling rate, which could result in the loss of crucial information.

    \item \textbf{Procedure}: 
    Participants wore a watch-like device with an IMU sensor on their non-dominant hand to collect data. During each session, participants were prompted to perform the target gesture through a chime or vibration, while an experimenter recorded the start and stop times of each gesture.
    To ensure an accurate representation of real user behavior, randomization was implemented throughout the study. A total of 100 sessions were conducted, with each session including all participants performing 50 instances of a single gesture recurrence for each of the four gestures. These gestures were paired with with 25 contexts, and each of the contexts mimics a real-life context.
    Examples of the 25 contexts are: \textit{standing/walking with hand up to the chest level/ with hand hanging, riding in a car a passenger, pushing stroller/cart with the other hand, holding a cup or something else in the other hand, sitting at the desk with elbow on the desk and hand in the air/ with arm laying on the desk/ with the elbow on the arm rest and hand in the air/ with the arm on an arm rest, laying down on back/side, lounging on the sofa (horizontal/slouched posture) with hand laying on the sofa or laps, cuddling/arm around someone else, stationary biking, walking up and down stairs, jogging, leaning over, picking something up}.\footnote{Table 1 in the Appendix shows the full list of contexts.}
    % Please see the full list of contexts in Table~\ref{tab:contexts} placed at the end of the paper.
    The contexts are also differentiated by if the participant is looking at the device or not.
    We also collected negative (i.e., non-gesture) examples under various activities such as jogging, running, biking, clapping, driving, waving, cleaning, etc. These examples are similar to the negative examples from~\cite{Xu2022Enabling}.
    This methodology aimed to capture the natural variability in users' motions without introducing any repetition bias.
\end{s_enumerate}

\section{Design Engineering Approach}
\label{sec:design_engineering}

Conventional machine learning approaches optimize the recognition model to best fit the data once. 
In an OWGR process, as illustrated in Figure~\ref{fig:process}, we additionally must optimize the policies by which the recognition model continually updates itself.
In the \emph{open-world gesture recognition} process, we outline five stages: (1) device deployment; (2) data logging; (3) data curation; (4) retraining trigger; and (5) updating policy. The final two stages highlight the two crucial design considerations that are integral and critical to an OWGR process that allows offline analysis:
\begin{s_enumerate}
    % \item \textit{With what data to train the recognition model?}
    % For each real-world application, how many samples should be sufficient to enable effective training for each task?
    \item \textit{When do we update the recognition model?} 
    How do we define the start and stop of a batch, and how frequently should we perform model updates?
    \item \textit{How do we update the recognition model?} 
    Which continual learning method provides the ideal trade-off between plasticity and stability, while minimizing processing and memory requirements?
    % Furthermore, how will the data collection process affect the final performance of the OWGR system? 
\end{s_enumerate}
% These include the three design questions proposed above.

We tackle these two considerations by adapting a design engineering approach~\cite{Kristensson2020Design,kristensson2021design} to perform a systematic evaluation of an OWGR process without long-term deployment, including user studies over multiple sessions, model deployment and run-time optimization, continuous data logging and collection, and so on. 
Further, the evaluation of an OWGR process involves system-level parameters beyond the recognition model hyper-parameters.

\begin{figure*}[t]
  \centering
  \begin{subfigure}[b]{0.9\textwidth}
  \centering
      \includegraphics[width=0.8\textwidth]{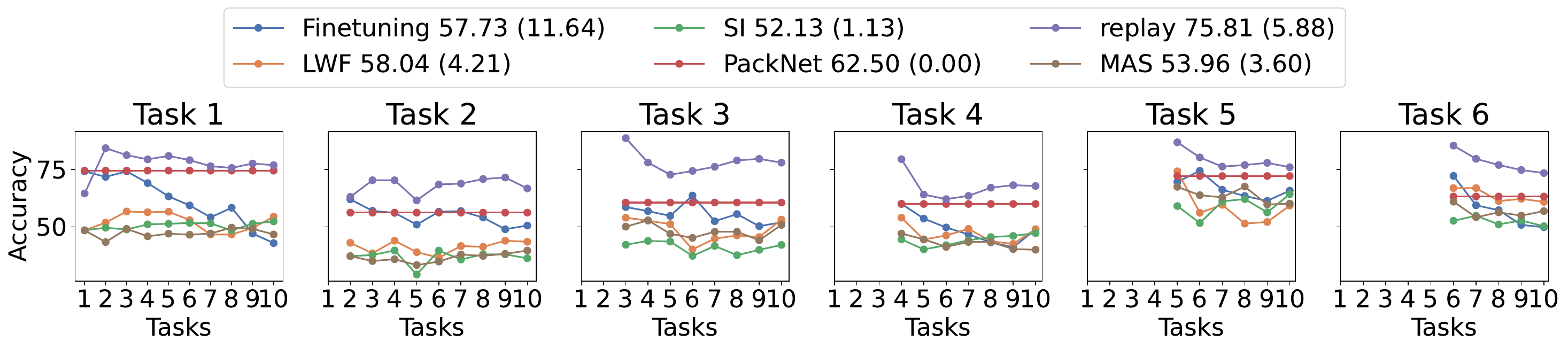}
    \caption{\emph{New Context}: Each new task requires one user to perform the same set of gestures under new context, such as standing, walking, riding in a car, laying down, etc. In this context, we have 10 tasks in total, due to size restrictions in presentation, we only show results up to Task 6.}
    \label{fig:new_context_demo}
  \end{subfigure}
  \hfill % optional: add space between images
  \begin{subfigure}[b]{0.9\textwidth}
  \centering
    \includegraphics[width=0.8\textwidth]{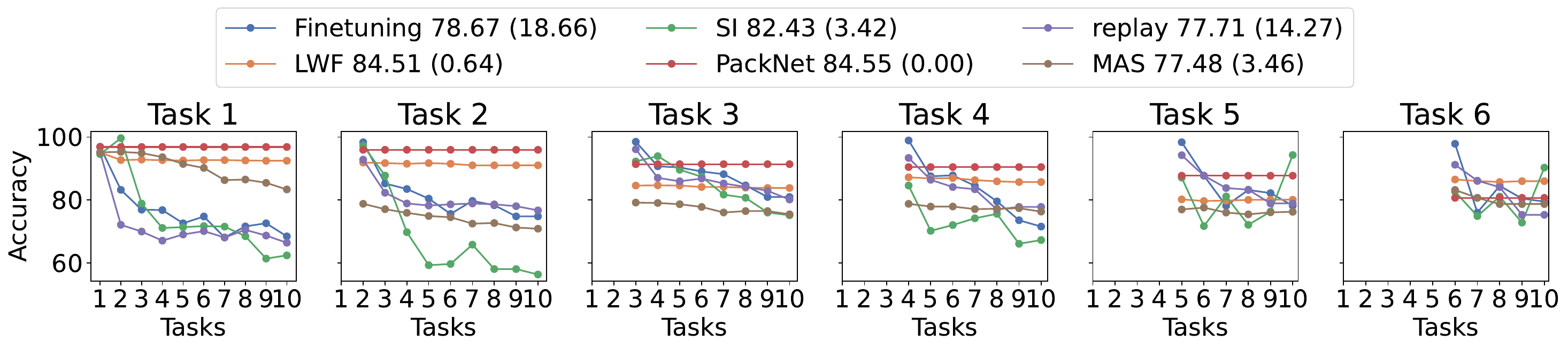}
    \caption{\emph{New User}: Each new task inquires 6 new users to perform the same set of gestures in the same context. This is akin to a model learning to generalize across different users' behavior patterns.}
    \label{fig:vr_poke_study}
  \end{subfigure}
  \caption{We examine the accuracy of a gesture recognition model, in an \emph{open-world gesture recognition} problem, tested on Task $n$, while the model is trained incrementally from Task $n$ to Task $N$. $N$ is the total number of tasks, and the model is trained by five continual learning methods and one baseline method: Finetuning. 
  For instance, in a subplot with title Task 2, we test the model, progressively trained from Task 2 through Task 10, exclusively on Task 2 and report 9 data points. 
  It lacks the data points from the model trained on Task 1 and tested on Task 2, because the tasks were introduced sequentially; a model trained on Task 1 has not been exposed to Task 2. 
  Consequently, the subplot for Task 2 omits these results. 
  % for a model trained solely on Task 1 and evaluated on Task 2. 
  This pattern continues with Task 3 and subsequent tasks, with a subplot of Task $n$ only reports $N-n+1$ data points (per method).
  The observed deep decline in accuracy from Finetuning in each subplot can be attributed to the model's tendency to forget earlier tasks as it learns new ones, which is referred to catastrophic forgetting.
  We proposed five continual learning methods to address this catastrophic forgetting problem in \emph{open-world gesture recognition}, namely SI, replay, LWF, PackNet and MAS, which are introduced in detail in Section~\ref{sec:function_carriers}. 
  In this Figure, we illustrate the results from performing preliminary experiments on two use cases of \emph{open-world gesture recognition}. We report average accuracy (forgetting) in the legend. Forgetting is the measure of the decrease in accuracy. We discuss these two measures in detail in Section~\ref{sec:metrics}.
  }
  \label{fig:new_demo}
\end{figure*}

\subsection{Functional Design and Function Carriers of OWGR}
\label{sec:function_carriers}

Following prior work~\cite{Kristensson2020Design,kristensson2021design}, we begin by identifying the main function (\texttt{Open-World Gesture Recognition}) and its key sub-functions (\texttt{Data Logging, Data Curation, Retraining Trigger, Updating Policy, Device Deployment}), along with their interrelationships.
% as shown in Figure~\ref{fig:process}.

Having established a function model, we then seek function carriers, or solution principles, and specifically for the \texttt{Updating Policy} function. We evaluate potential function carriers by considering their controllable and uncontrollable parameters through envelope analyses that simulate emergent outcomes as a function of the controllable and uncontrollable parameters in the system. This helps us understand the system requirements for these function carriers to effectively fulfill their functions.

Our study mainly examines two functions, \texttt{Retraining Trigger} and \texttt{Updating Policy}, which both require dedicated online analysis studies. 
This approach aids in quantitative parameter investigation, thereby enhancing our understanding of continual learning methods for \emph{open-world gesture recognition}. A comprehensive functional design of the entire system is beyond this paper's scope.

% \subsubsection{Continual Learning Methods}
% \label{sec:continual_function}
We here study six function carriers. 
The study of these function carriers will determine which continual learning method to use to update the gesture recognition model.
By varying the function carrier, we can explore the design question: \emph{how do we update the recognition model?}.
We evaluate the following five different continual learning methods, and a baseline, across the aforementioned families in Section~\ref{sec:continual_learning}: 
% (we only introduce the central idea of the methods below, for detailed implementation, please refer to the original papers):
\begin{s_enumerate}
    \item \textbf{Baseline (finetuning)}: For baseline, we use naive finetuning, which optimizes the model trained from previous task on the current task. This method greedily trains each task without considering previous task performance~\cite{delange2021continual}. 
    \item \textbf{Learning without Forgetting (LwF)}: Learning without Forgetting (LwF)~\cite{li2017learning} retains knowledge of preceding tasks by means of knowledge distillation.
    \item \textbf{Synaptic Intelligence (SI)}: In SI~\cite{zenke2017continual}, each synapse accumulates task-relevant information over time and exploits this information to rapidly store new memories without forgetting old ones~\cite{zenke2017continual}.
    \item \textbf{PackNet}: PackNet~\cite{mallya2018packnet} iteratively assigns parameter subsets to consecutive tasks by constituting binary masks. For this purpose, new tasks establish two training phases. The first phase is training the recognition model without altering previous task parameter subsets and then pruning a portion of unimportant free parameters. The second training phase retrains the remaining subset of important subsets. Therefore, PackNet allows explicit allocation of network capacity per task, and therefore inherently supports zero forgetting on previous tasks. However, the disadvantage of PackNet is that the total number of tasks is limited.
    \item \textbf{Replay}: We use finetuning with an arbitrary replay buffer, which exploits available exemplar memory up to the replay buffer size and incrementally divides equal memory capacity over all previous tasks~\cite{delange2021continual}. 
    \item \textbf{Memory Aware Synapses (MAS)}: MAS~\cite{aljundi2018memory} computes the importance of the parameters of a neural network in an unsupervised and online manner. When learning a new task, changes to important parameters can be penalized by the accumulated importance measure for each parameter of the network, effectively preventing important knowledge related to previous tasks from being overwritten~\cite{aljundi2018memory}.
\end{s_enumerate}
Both SI and MAS are parameter prior-based regularization methods, whereas LwF is a data prior-based regularization method. MAS is a parameter isolation method.

\changetextt{
Additionally, we carried out a preliminary offline experiment with internal participants who were asked to perform gestures under two distinct scenarios. 
In this experiment, we tested the above six different continual learning methods, aiming to adapt the gesture recognition model.
% including fine-tuning.
% and five other approaches that will be discussed in detail in Section~\ref{sec:function_carriers}, aiming to adapt the gesture recognition model. 
The outcomes of this study are presented in Figure~\ref{fig:new_demo}. This preliminary analysis primarily highlights three key findings:
\begin{s_itemize}
\item Naive fine-tuning with new task data leads to significant forgetting of previously learned tasks, underscoring the necessity for continual learning methods to address this issue. This emphasizes the critical nature of the \emph{open-world gesture recognition} paradigm.
\item Continual learning methods demonstrate varied effectiveness across different scenarios; a method that shows superior performance in one case may not be the best choice for another. This underlines the importance of conducting comprehensive quantitative analyses on large-scale \emph{open-world gesture recognition} datasets to provide design guidance.
\item The execution of this preliminary study required substantial resources and time, as participants needed to perform gestures in a variety of contexts and combine different gestures. This underscores the critical need for online experiments to supplement or replace such extensive offline analyses.
\end{s_itemize}
}

\subsection{Surrogate Task Model}

In envelope analysis, a surrogate model is a simplified variant of a complex, computationally involved model. It involves identifying and varying controllable and uncontrollable system parameters to evaluate a system with respect to its emerging outcomes. Controllable parameters can be fine-tuned for optimization, while uncontrollable ones help predict potential performance variations (sensitivity analysis)~\cite{Kristensson2020Design}. Envelope analysis is performed by altering one parameter at a time, with the other parameters held constant.

We here propose a surrogate task model in which we can vary different task settings.
The task setting in each case determines the specific definition of each task which is represented by the task descriptor $t_{i}$ in a triplet $(x_{i},y_{i},t_{i})$.
By varying this setting, we can answer the other design question: \emph{when do we update the recognition model?}, because the task setting defines the start and stop of a batch, the size of a batch, and the frequency of model updates, and so on.
% : \emph{when to update the recognition model}, 
% such as, \emph{How many tasks can each continual learning method learn without catastrophic forgetting?}
We have the following parameters in the surrogate task model that determines the task setting:
\begin{s_enumerate}
    \item \textit{Granularity of tasks}: The tasks can be either fine-grained or coarse. For example, in the case of \emph{new context}, coarse task granularity is \textit{standing} and fine task granularity is \textit{standing with hand up to the chest level}. 
    This parameter is controllable.
    \item \textit{Order of tasks}: By varying the order in which new tasks are presented (e.g. easier-to-harder vs harder-to-easier), we can explore whether the order of tasks affects the learning result. This parameter is uncontrollable because new tasks do not arrive in pre-defined order. 
    The easiness of a task is determined by the preliminary results obtained through testing the accuracy of a classification model on that task. A higher accuracy level indicates an easier task.
    \item \textit{Number of tasks}: This is equivalent to the number of tasks in total. In the case of \emph{new context}, by investigating the total number of contexts, we can observe the capacity of each continual learning method. This parameter is controllable because we set the maximum total number of tasks to which the system can adapt.
\end{s_enumerate}
These task settings are applied differently to each of the use cases:
\begin{s_enumerate}
    \item \emph{New context}: We incorporate all parameters from the surrogate task model. 
    The default task setting is coarse task granularity provided in a random order. The default number of total tasks is set to 10 (as in \cite{delange2021continual}).
    \item \emph{New user}: We only include the controllable parameter \textit{number of tasks}.
    Here the number of tasks is the number of total users.
    The default setting is 15 users, selected at random. 
\end{s_enumerate}

\subsection{Evaluation Metrics}
\label{sec:metrics}
We use accuracy and forgetting as the evaluation measures for the OWGR model, with definitions adopted from prior work~\cite{chaudhry2018riemannian} as follows:
\begin{s_enumerate}
    \item \textbf{Accuracy}: 
    Let $a_{k,j} \in \left [ 0,1 \right ]$ be the gesture recognition accuracy (fraction of correctly classified gesture events) evaluated on the held-out test set of the $j$-th task ($j \leq  k$) after training the network incrementally from tasks 1 to $k$. The accuracy measure at task $k$ is then defined as $A_{k} =\frac{1}{k} \sum_{j=1}^{k}a_{k,j}$.
    The average accuracy measures the plasticity of the system in transferring knowledge from old tasks to new tasks.
    % The higher the accuracy, the better the recognition model. 
    \item \textbf{Forgetting}: 
    We define forgetting for a particular task as the difference between the maximum knowledge gained about the task throughout the learning process in the past and the knowledge the model currently has about it. This, in turn, gives an estimate of how much the model forgot about the task given its current state (catastrophic forgetting).
    Following this, we quantify forgetting for the $j_{th}$ task after the mode has been incrementally trained up to task $k > j$ as: $f_{j}^{k} = \underset{l \in \left \{ 1,.., k-1 \right \}}{\mathbb{E}}a_{l,j}-a_{k,j}, \forall j<k$.
    % \begin{equation}
    %     f_{j}^{k} = \underset{l \in \left \{ 1,.., k-1 \right \}}{\mathbb{E}}a_{l,j}-a_{k,j}, \forall j<k
    % \end{equation}    
    Note, $f_{j}^{k} \in \left [ -1,1 \right ]$ is defined for $j<k$ as we are interested in quantifying forgetting for \textit{previous} tasks.
    Moreover, by normalizing against the number of tasks seen previously, the average forgetting at $k-th$ task is written as $F_{k} = \frac{1}{k-1} \sum_{j=1}^{k-1}f_{k,j}$. Lower $F_{k}$ implies less forgetting on previous tasks. 
\end{s_enumerate}

% \hspace{4em}
% \hfill
In this paper, we do not report accuracy using different metrics, such as F1 scores, because the contribution is not to propose state-of-the-art gesture recognition models.
Instead, the focus is on comparing the relative accuracy and the forgetting measures between different continual learning methods to provide design guidance for developing \emph{open-world gesture recognition} methods. 
Furthermore, given the extensive experiments conducted under various settings in this paper, it is impractical to include and report on different metrics. 
Therefore, we report on the final accuracy and forgetting, obtained by evaluating each task after learning the entire task sequence.

% \begin{figure}[t]
% \centering
%   \includegraphics[width=0.6\linewidth]{figures/quartznet.pdf}
%   \caption{QuartzNet architecture for our gesture recognition model.}
%   \label{fig:quartznet}
% \end{figure}

\begin{figure*}[t]
\centering
  \includegraphics[width=0.7\linewidth]{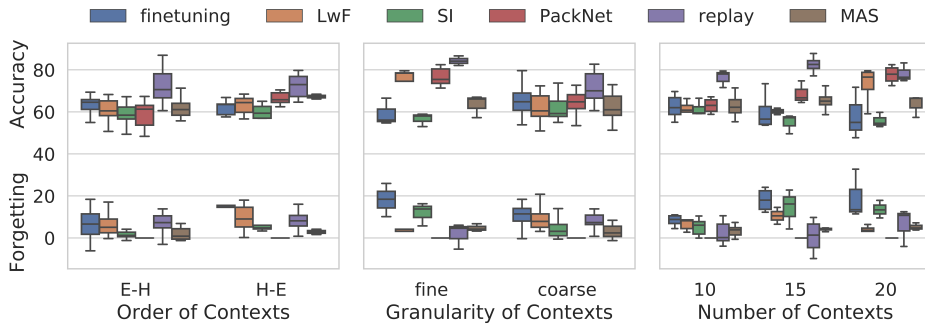}
  \caption{\textbf{Effect of Task Setting on Performance for \emph{New Context}} 
  Each sub-figure represents one task setting.
  Each box plot shows mean, median, and quartile data.
  a) \textbf{Order of Contexts}: Change of context order does not significantly affect average accuracy and forgetting measure for most methods. The X-axis represents a specific order of the tasks. E-H is easy-to-hard, H-E is hard-to-easy. b) \textbf{Granularity of Contexts}: Some methods perform particularly better or worse in fine-grained context than coarse context. The X-axis represents that if the context is coarse or fine. c) \textbf{Number of Contexts}: Different methods exhibits various performances when number of contexts (tasks) increases. The X-axis represents the total number of contexts (tasks)}. 
  \label{fig:new_context_task_function}
\end{figure*}

\begin{figure}[t]
\centering 
    \includegraphics[width=0.8\textwidth]{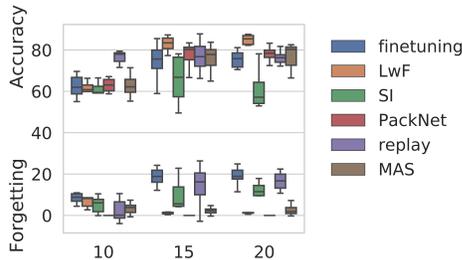}
    \caption{\textbf{Effect of Number of Total Tasks for \emph{New User}.} 
      A larger number of total users leads to higher accuracy and forgetting measure for most methods. Each box plot shows mean, median, and quartile data.
      The X-axis represents the total number of users that the model must learn in the \emph{New User} case.}
    \label{fig:new_user_task_size}
\end{figure}

\subsection{Implementation Details}

We implement the continual learning algorithms based on an open-source generalizing continual learning framework~\cite{delange2021continual} in PyTorch. For data processing, we use a sliding window approach to segment the data for both training data and testing data preparation, we use a window size of 120 and a window step size of 60. 
We adopt the same parameters for debouncing thresholds in~\cite{Shen2022Gesture}.
For the hyperparameter search of each continual learning method, we conduct an approach that first tries to decay each hyperparameter separately, and decays all if none of these individual decays to achieve accuracy within the finetuning margin. This process is then repeated until the stability decay criterion is met
For the specific setup of the training framework, we define maximal plasticity search with a coarse learning rate grid $\left \{1e^{-2}, 5e^{-3}, 1e^{-3}, 5e^{-4}, 1e^{-4}\right \}$. We set the finetuning accuracy drop margin to 0.2 and the decaying factor to 0.5. We use a Stochastic Gradient Descent with a momentum of 0.9 and batch size of 128. We use a max of 100 training epochs with early stopping and annealing of the learning rate. That is the learning rate decays with factor 10 after 10 unimproved iterations of the validation accuracy and the training process should terminate after 15 unimproved iterations of training.

We use a simplified version of QuartzNet architecture~\cite{kriman2020quartznet}, a state-of-the-art convolutional architecture for speech recognition, for our gesture recognition model described by $f$ in Equation~\ref{eq:risk1}. 
It is used as the classifier that takes IMU signals as input and predicts among the gesture classes.
The selection of the QuartzNet model is based on preliminary experiments.
\changetextt{
We began with a literature review on state-of-the-art classification models for time series data and tested over 10 models.
Our experiments revealed that the QuartzNet model achieves state-of-the-art performance in real-time gesture recognition, striking an optimal balance between accuracy and latency.
We further improved the performance by hyperparameter sweeping making it suitable for IMU data classification.
}
This is attributed to the model's quantization, which allows it to be compact enough to operate in real-time on smaller devices.
To make the model deployable on more memory-constrained devices, we have removed the residual connections, grouped point-wise convolutions, and channel shuffle without scarifying the accuracy performance.
\footnote{Figure 2 in the Appendix illustrates the details of the model architecture.}

\section{Results}
This section describes the results of varying the surrogate task model for each of the two use cases.

% \subsection{New Context}
Figure \ref{fig:new_context_task_function} consists of three sub-figures which describe the effect of different task settings separately for the \emph{new context} case.
Prior work suggests that an easy-to-hard task ordering might achieve better performance than a hard-to-easy ordering~\cite{wang2021survey}. 
However, Figure \ref{fig:new_context_task_function} a) shows that the impact of the task order on the average accuracy and forgetting measure is insignificant.
We do observe that the variance of the performance scores is considerably smaller for easy-to-hard ordering. 
In Figure \ref{fig:new_context_task_function} b), when setting the granularity of contexts to fine-grained as the task-setting, we see that finetuning shows a lower accuracy and a higher forgetting measure, whereas methods such as LWF, PackNet, and replay reach a higher accuracy while maintaining a low forgetting measure. 
A coarse context setting has more training data, as a coarse context consists of multiple fine-grained contexts.
This suggests that these continual learning methods can successfully optimize plasticity (transfer knowledge of previous tasks to new tasks) to achieve data-efficiency. 
Figure \ref{fig:new_context_task_function} c) shows that a lower number of total tasks can stabilize the average performance of the continual learning methods. 
A larger number of total tasks leads to an increased forgetting measure as the model's capacity for learning new tasks is limited. 
However, PackNet and LwF can successfully re-use the previous knowledge to produce an increased accuracy on new tasks. 
On the other hand, finetuning introduces a significant forgetting measure accompanied by a low accuracy score when the number of tasks is large.

For the \emph{new user} case, we observe from Figure \ref{fig:new_user_task_size} that a larger number of users increases both accuracy and forgetting metrics with finetuning and replay methods.
In contrast, most continual learning methods such as PackNet, MAS and LWF achieve high accuracy while maintaining low forgetting measure with larger number of total users. 
This suggests that, in the \emph{new user} case, PackNet, MAS, and LwF can better transfer knowledge of old users to new users.

% \section{Validation and Evaluation}

\section{Discussion}
% In this section, we will discuss three key areas.
% : \textit{How ?}, \textit{When ?} and \textit{What ?}.

\subsection{Balance between Stability and Plasticity}

\changetextxx{
Continual learning aims to strike a balance between two key objectives: stability and plasticity. Stability refers to the ability to retain previously learned knowledge over time, avoiding catastrophic forgetting. Plasticity, on the other hand, refers to the ability to acquire new knowledge from new data distributions, avoiding overfitting to old data. Stability is typically measured by evaluating the degree of forgetting or performance degradation on previously learned tasks or data distributions. Plasticity is measured by assessing the model's accuracy or performance on new, unseen data distributions.
When designing OWGR processes, it is crucial to consider this stability-plasticity trade-off. Design choices and techniques should be evaluated based on their ability to achieve an appropriate balance between these two objectives.
}

\subsection{Design Guidelines from Envelope Analysis}
% We now revisit the two general design questions we proposed in Section \ref{sec:design_engineering}.

\subsubsection{When do we update the recognition model?}
\label{sec:when}

\changetextxx{This question asks for the best task setting for each case. This involves defining the boundaries between tasks and determining when a new task has commenced. Once a new task is detected, a retraining trigger (Figure 2) activates the update of the recognition model using a continual learning method. Thus, we use the results from the envelope analysis from the surrogate task model to answer this question, and we explore which task settings yield the most favorable outcomes. }
Most of the parameters in the surrogate task model are controllable by the designers.
\textbf{We advise that a coarse context usually gives better results for the \emph{new context} case.}
A coarse context is also more practical, as it simplifies the activity recognition needed to produce a task descriptor.
\changetextxx{The accuracy of a task descriptor determines the granularity of tasks; a less accurate descriptor results in a coarser context, while a more precise descriptor leads to a finer-grained context. We previously discussed that the \emph{new context} case may require classification of the current activity context, which can be performed by AR/VR HMDs and wearables via activity detection through different sensors~\cite{Cornacchia2017ASO, Ma2016LearningAP}. The precision of state-of-the-art activity classifiers, acting as task descriptors, is sufficiently high that it does not adversely affect the primary gesture recognition model. Therefore, these task descriptors are generalizable descriptors and do not require re-training with new real-world data. }
\textbf{For the \emph{new user} case, we see the best results when the number of different users is large with LwF and PackNet}, as we observe that a larger number of users increases the overall accuracy for these two methods. 
\textbf{In practice, a model needs to be pre-trained and it is encouraged to pre-train the model on a large-scale dataset if available.}

\subsubsection{How do we update the recognition model?}
% First, we answer the question of \textit{how to update the recognition model}.
We observe that naive fine-tuning experiences serious catastrophic forgetting and is not suitable for OWGR. 
For the \emph{new context} case, the replay method outperforms other continual learning methods by a large margin in accuracy.
% The buffer size should be set as large as possible under the consideration of memory efficiency, as we did not observe any significant performance degradation with overly large buffer size. 
% The empirical results also eliminate the concern of negative impact on accuracy caused by an overly large buffer size.
In contrast, PackNet has zero forgetting by its design and still maintains relatively high accuracy.
\textbf{Overall, for the \emph{new context} case, the replay method is the best choice to optimize accuracy---memory/privacy constraints permitting---whereas PackNet is a better choice to balance accuracy and forgetting, provided the model capacity is sufficient for the total number of tasks.} 
For \emph{new user}, the replay method no longer ranked as the top performer in accuracy. 
Both PackNet and LwF exhibit more stable performance in this case as the difference between the first quartile (Q1) and the third quartile (Q3) is smaller, showing a smaller variance in the performance metrics. 
\textbf{Thus PackNet and LwF is a better choice for the \emph{new user} case.}

\subsection{Envelope Analysis: A Practical Alternative to Large-Scale User Studies in OWGR}

The primary objective of this research is to introduce an novel method for assessing continual learning methods for OWGR, bypassing the need for large-scale user studies. Researchers often lack resources for such studies, which are more demanding than evaluating conventional gesture recognition models in controlled settings. The proposed approach uses envelope analysis with parameterization of key functions, offering a robust and practical alternative to user studies. By varying parameters, different scenarios can be systematically explored. The extensive and diverse dataset used in this research provides a strong foundation for evaluating the OWGR process, reducing the need for a separate large-scale user study.
Although some conclusions drawn from the envelope analysis might appear intuitive, there has been a notable absence of such thorough experiments in the realm of what we would classify as OWGR. Further, assumptions from other, related, areas might not directly translate to gesture recognition problems relevant for mixed reality applications.

\subsection{OWGR in Mixed Reality}
\changetextxx{
OWGR enables gesture recognition models to adaptively learn new gesture data patterns from diverse contexts and users on-the-fly, eliminating the need for extensive pre-deployment data collection. It facilitates personalization by learning and adapting to individual users' contexts, whether performed in static or dynamic environments. For shared devices among multiple users, such as families, it eliminates recalibration needs by remembering previous users' patterns while continuously learning from new user. This ability to seamlessly learn and retain knowledge across contexts and users ensures a robust, calibration-free experience.
}

\subsection{Limitation and Future Work}
\label{sec:limitations}

One limitation in our study is the lower accuracy in gesture recognition. 
It's worth noting that the accuracy reported in this paper represents the final model's accuracy on the last task, following the entire task sequence (see Section~\ref{sec:metrics}). The accuracy diminished from the initial task to the final task due to catastrophic forgetting. This further encourages us to refine our continual learning algorithms as future work. 
Another contributing factor is that our training algorithm wasn't explicitly tailored for optimal accuracy.
Techniques such as data augmentation, feature engineering, and model pre-training, which are commonly used to enhance model performance~\cite{Xu2022Enabling,Shen2022Gesture,Shen2021Imaginative}, were not employed in our study. This decision was intentional, as our primary aim was to isolate the effects of the envelope analysis. By excluding these optimization techniques, we ensured that any observed effects on the final results were solely due to changes in the parameters of the envelope analysis.

\section{Conclusion}
% A key challenge in mixed reality is enabling seamless, accurate gesture recognition. 
This paper presents a design engineering approach, using a large-scale dataset in combination with envelope analysis, to explore continual learning methods for the \emph{open-world gesture recognition} process. 
% This approach is applicable even when training and testing data do not share similar characteristics. 
We discuss the significant catastrophic forgetting observed in the baseline fine-tuning approach and examine the positive and negative qualities of various continual learning methods.
To assist other researchers and developers we present guidelines derived from the extensive data collection and the computational experiments. Intended for developers and designers of wrist-worn gesture-sensing systems, our guidelines provide an alternative to traditional large-scale user studies, lowering the barriers and promoting a more refined OWGR process. This paper is also a demonstration of how a design engineering methodology, in particular envelope analysis, offers a new route for system evaluations in this area, potentially inspiring other researchers in mixed reality to adopt similar methods for real-world assessment challenges.

\bibliographystyle{abbrv-doi}

\bibliography{template}
\end{document}